%% file: main.tex
\documentclass[letterpaper, 10 pt, conference]{ieeeconf}
\IEEEoverridecommandlockouts 
\input{math_commands.tex}

\usepackage{cite}
\usepackage{amsmath,amssymb,amsfonts}
\usepackage{algorithmic}
\usepackage{graphicx}
\usepackage{textcomp}
\usepackage{xcolor}
\usepackage{hyperref}
\usepackage{url}
\usepackage[linesnumbered,ruled,vlined]{algorithm2e}
\usepackage{adjustbox}
\usepackage{tabu}
\usepackage{booktabs}       
\usepackage{multirow}
\usepackage{multicol}
\usepackage{rotating}
\usepackage{subfig}
\usepackage{dsfont}

\def\BibTeX{{\rm B\kern-.05em{\sc i\kern-.025em b}\kern-.08em
    T\kern-.1667em\lower.7ex\hbox{E}\kern-.125emX}}
\begin{document}

\title{\LARGE\bf Guided Online Distillation: \\ Promoting Safe Reinforcement Learning by Offline Demonstration
    \thanks{
        \textsuperscript{1} J. Li, B. Zhu, J. Jiao, M. Tomizuka, C. Tang, and W. Zhan are with the University of California, Berkeley, CA, USA. 
        Correspondence to: Chen Tang \texttt{<chen\_tang@berkeley.edu>}.
    }
    \thanks{
        \textsuperscript{2} X. Liu is with the University of Michigan, Ann Arbor, MI, USA.
    }
}
\author{
Jinning Li\textsuperscript{1}
\and
Xinyi Liu\textsuperscript{2}
\and
Banghua Zhu\textsuperscript{1}
\and
Jiantao Jiao\textsuperscript{1}
\and
Masayoshi Tomizuka\textsuperscript{1}
\and
Chen Tang\textsuperscript{1}
\and 
Wei Zhan\textsuperscript{1}
}

\maketitle
\thispagestyle{empty}
\pagestyle{empty}

\begin{abstract}
Safe Reinforcement Learning (RL) aims to find a policy that achieves high rewards while satisfying cost constraints. When learning from scratch, safe RL agents tend to be overly conservative, which impedes exploration and restrains the overall performance. 
In many realistic tasks, e.g. autonomous driving, large-scale expert demonstration data are available.
We argue that extracting expert policy from offline data to guide online exploration is a promising solution to mitigate the conserveness issue.
Large-capacity models, e.g. decision transformers (DT), have been proven to be competent in offline policy learning.
However, data collected in real-world scenarios rarely contain dangerous cases (e.g., collisions), which makes it prohibitive for the policies to learn safety concepts.
Besides, these bulk policy networks cannot meet the computation speed requirements at inference time on real-world tasks such as autonomous driving. To this end, we propose Guided Online Distillation (GOLD), an offline-to-online safe RL framework. GOLD distills an offline DT policy into a lightweight policy network through guided online safe RL training, which outperforms both the offline DT policy and online safe RL algorithms. Experiments in both benchmark safe RL tasks and real-world driving tasks based on the Waymo Open Motion Dataset (WOMD)~\cite{Ettinger_2021_ICCV} demonstrate that GOLD can successfully distill lightweight policies and solve decision-making problems in challenging safety-critical scenarios. 
\end{abstract}

\section{Introduction}

Safe Reinforcement Learning (RL) aims to find a policy that not only achieves high rewards but also keeps the cost of violating constraints below a specified threshold. 
Traditional online safe RL algorithms~\cite{achiam2017constrained, liu2022constrained, li2021safe} solve for an optimal safe policy by performing online rollouts in an environment and updating the policy accordingly. 
However, these algorithms always start training policies from scratch. 
The agent needs to learn to locate and avoid hazardous areas while it is still struggling to discover high rewards in the environment.
The safety constraints discourage the agent from exploring certain hazardous areas~\cite{shperberg2023relaxed}, which leads to a pitfall that induces the policy to be overly conservative.
The overly conservative policy often causes the agent to get stuck during its exploration, surrounded by complex hazard areas.
Jammed at some states repetitively causes a skewed data distribution in the replay buffer, which deceives the policy that these states are the highest possible reward areas. It thus impedes the learning process and the overall performance. 

In this situation, a near-optimal policy extracted from offline demonstrations can serve as a guide during online fine-tuning.
Jump Start Reinforcement Learning (JSRL)~\cite{uchendu2023jump}, as an online fine-tuning method, has proven that training a new policy for online adaptation while using an offline extracted guide policy can be effective in regular RL settings, compared to naively initializing RL by the pre-trained policy~\cite{uchendu2023jump}.
It is natural and intuitive to propagate this meta-training scheme to the safe RL domain.
The guide policy helps the agent being trained online start exploration from high-reward areas, and build new skills based on it thereafter.
It is promising to save the agent from getting stuck in hazardous areas during exploration.
Therefore, we propose to adapt JSRL to the safe RL setting, so that a better reward-cost trade-off can be achieved in those application scenarios where offline demonstration is available. 

In many real-world situations, large-scale datasets already exist that can provide expert demonstrations for training policies\cite{Ettinger_2021_ICCV, Kan_2023_arxiv, caesar2020nuscenes, chang2019argoverse, li2021spatio, wu2022prim}.
Prior work on imitation learning~\cite{hussein2017imitation, codevilla2018end} and offline RL~\cite{kostrikov2021offline, li2022hierarchical, li2022dealing} has investigated extracting high-performance policies directly from offline datasets to avoid risky online exploration or learning by trial and error.
While it is seemingly promising to extract near-optimal policies with high rewards from offline datasets, prevalent Behavior Cloning (BC) or offline RL algorithms~\cite{mandlekar2021matters, kumar2022should} tend to fail when the demonstrations come from human experts. 
Decision transformer (DT)~\cite{chen2021decision} has been shown as a strong method in such settings compared to these algorithms.
It adopts large-scale models that are proven to have potentials~\cite{vaswani2017attention, brown2020language, ramesh2021zero}.
Therefore, we explore the possibility of applying high-capacity decision transformers to learn from offline expert demonstrations in this paper.
However, easily accessible datasets often lack sufficient data points in safety-critical scenarios, such as collisions in real-world traffic datasets~\cite{ding2020cmts, li2020interaction}.
Consequently, offline datasets alone cannot provide enough information on the safety constraints in the environment, and thus offline training is not sufficient for safe RL.
It, therefore, strengthened the necessity of continuing to improve the decision-making policy by an online finetuning process with interactions in task environments~\cite{nair2020awac, nakamoto2023cal}.

Prior work~\cite{nair2020awac, nakamoto2023cal, zhang2023efficient, wang2022offline, sun2021online} typically uses the offline trained policy network architecture for online finetuning. 
Unfortunately, DT's transformer-based policy network, with its numerous parameters, can often fall short of meeting computation speed requirements in real-world tasks like autonomous driving.
However, we do not intend to directly shrink the network size in offline training because it will sacrifice its performance to a great extent, and our experiments show that the performance and efficiency of online finetuning largely rely on the quality of the offline trained guide/expert policy. 
Alternatively, we sought to distill a more computationally efficient policy from DT during online training. 
There are many prior works on network distillation~\cite{rusu2015policy, czarnecki2019distilling, traore2019discorl}, but their student policies are trained either by supervised learning to replicate the teacher's behavior, which isolates the student policy and the environment forbidding active explorations or by indirect ways such as reusing the critic of the teacher, which does not fully incorporate the extracted prior skills from offline demonstrations. 
Training a different policy with the guidance of DT instead of initializing RL with existing policy in JSRL~\cite{uchendu2023jump} allows us to change the policy network architecture and encourage the agent to explore more promising areas, which unifies the purpose of finetuning and distillation in this paper.

In summary, we propose a training scheme, named Guided Online Distillation (GOLD), for safe RL tasks where offline expert demonstration is available. GOLD leverages an offline learned large-scale policy to guide the online learning of a computationally efficient, safe RL policy. Compared to safe RL from scratch, GOLD can improve the cumulative reward achieved by the policy while maintaining the cumulative cost below the threshold. In summary, our contributions include:

\begin{itemize}
    \item We propose a training scheme, Guided Online Distillation (GOLD), for safety-critical scenarios where offline expert demonstration is available. It solves the problem caused by limited high-risk cases in offline datasets and conservative exploration in safe RL.
    \item We empirically show that adopting a DT instead of BC improves the performance of the offline extracted policy, and the large-capacity and well-performed DT guide policy is crucial for the online distilled lightweight policy to optimize its reward-cost trade-off. 
    \item We train and evaluate the proposed algorithm on both benchmark safe reinforcement learning and real-world autonomous driving tasks extracted from the Waymo Open Motion Dataset (WOMD)~\cite{Ettinger_2021_ICCV, Kan_2023_arxiv}. We show that GOLD can effectively accelerate online learning and improve policy performance.
\end{itemize}


\section{Preliminaries}

\subsection{Constrained Markov Decision Process}

We define a Constrained Markov Decision Process (CMDP) by a tuple $\mathcal{M}:=(\mathcal{S}, \mathcal{A}, \mathcal{P}, r, c, \gamma, \mu_0)$, where $\mathcal{S}$ is the state space, $\mathcal{A}$ is the action space, $\mathcal{P}: \mathcal{S} \times \mathcal{A} \times \mathcal{S} \rightarrow [0, 1]$ is the transition function specifying the probability $p(\vs_{t+1} | \vs_t, \va_t)$ from state $\vs_t$ to $\vs_{t+1}$ when applying $\va_t$, $r: \mathcal{S} \times \mathcal{A} \rightarrow \R$ is the reward function, $c : \mathcal{S} \times \mathcal{A} \rightarrow [0, C_m]$ is the cost function for violating the constraint with $C_m$ as the maximum cost~\cite{altman1998constrained}, $\gamma$ is the discount factor, and $\mu_0: \mathcal{S} \rightarrow [0, 1]$ is the initial state distribution. 

In safe RL, the goal is to find a policy $\pi \in \Pi$ where $\Pi$ is the policy class such that it obtains a high return in reward and maintains the cost return below a threshold $\kappa \in \R^+$. 
Formally, we denote the reward value function $V_r^\pi(\mu_0) = \E_{\tau \sim \pi,\vs_0 \in \mu_0} [\sum_{t=0}^\infty \gamma^t r(\vs_t, \va_t)]$ as the discounted cumulative reward under the policy $\pi$ and the initial state distribution $\mu_0$, where $\tau = \{ \vs_0, \va_0, \dots \}$ is the trajectory.  
The cost value function is defined similarly as $V_c^\pi(\mu_0) = \E_{\tau \sim \pi,\vs_0 \in \mu_0} [\sum_{t=0}^\infty \gamma^t c(\vs_t, \va_t)]$. 
The objective is then to find an optimal policy $\pi^*$ by solving the following constrained optimization problem:
\begin{gather}
    \pi^* = \arg \max_\pi V_r^\pi(\mu_0), \: \text{s.t. } V_c^\pi(\mu_0) \leq \kappa.
    \label{eq:safe-rl-obj}
\end{gather}

\subsection{Reward-Cost Relationship}
\label{sec:reward-cost-relationship}

\begin{figure}[t]
    \centering
    \includegraphics[width=0.49\textwidth]{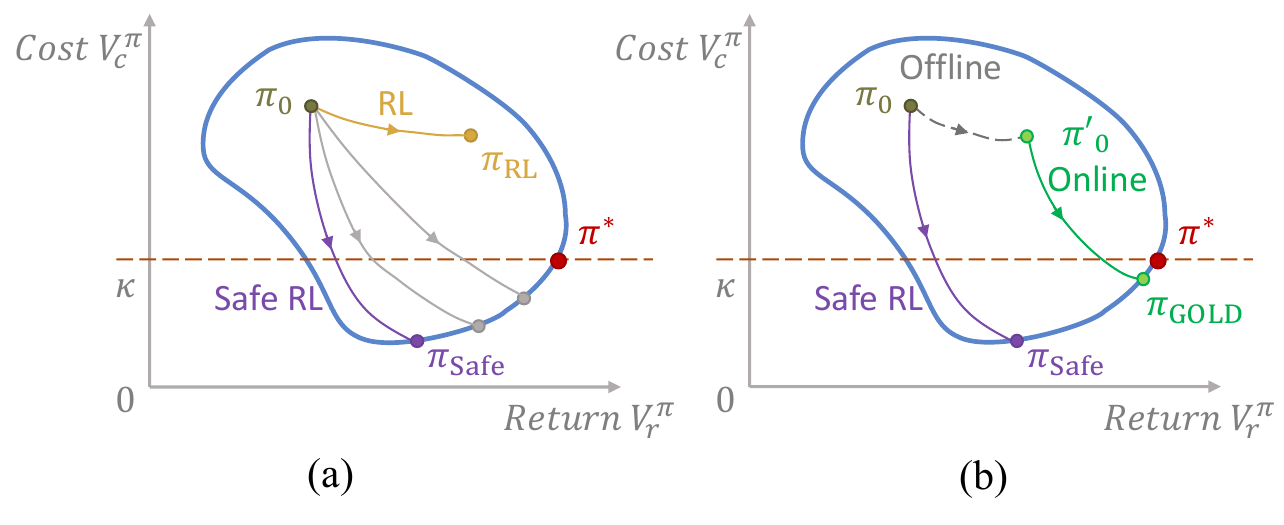}
    \caption{
        An illustration of reward-cost relation for all policies in a certain environment.
        (a) Regular and safe RL policies converge to points far away from the optimal policy $\pi^*$.
        (b) Trained from offline demonstration, $\pi'_0$ is attracted toward $\pi^*$, and hence our method results in a lightweight yet more capable policy during online distillation.
    }
    \label{fig:reward-cost}
\end{figure}

The constraint in Eqn.~\ref{eq:safe-rl-obj} illustrates that a safe RL algorithm needs to control two signals simultaneously, i.e., reward and cost, compared to regular RL.
A plot of cumulative cost and reward pair of policies on a $2d$-plane is informative for safe RL algorithm analysis.
Given a specific environment, each policy $\pi \in \Pi$ can be mapped onto a point within the blue circle, representing the reward-cost return pair $(V_r^\pi, V_c^\pi)$ that $\pi$ obtains in the environment~\cite{liu2022robustness}.
All policies that lie below the threshold $\kappa$ (the orange dash in Fig.~\ref{fig:reward-cost}) are considered feasible solutions.
The optimal policy $\pi^*$ obtains the highest possible reward return while maintaining the cost return below the threshold $\kappa$. When a policy is being trained in the environment, its corresponding reward-cost pair moves on this $2d$-plane.
Most RL algorithms randomly initialize a policy $\pi_0$, which places it on the upper left corner in Fig.~\ref{fig:reward-cost}. 
If being evaluated, $\pi_0$ will obtain low reward and high cost.

Assume RL algorithms are effective in the environment. 
For a regular RL algorithm, e.g., PPO \cite{schulman2017proximal} or SAC\cite{haarnoja2018sac}, this means the reward obtained by the policy continuously increases, but there is no consideration of the cost return. 
This results in the yellow trajectory in Fig.~\ref{fig:reward-cost}(a).
On the contrary, a safe RL algorithm~\cite{ladosz2022exploration} is dedicated to decreasing the cost and increasing reward simultaneously. 
Hence, its trajectory moves toward the bottom right corner in Fig.~\ref{fig:reward-cost}(a).

Once a trajectory reaches the boundary of the feasible area below $\kappa$, it stops moving because the safe RL does not allow an increase in cost or a decrease in reward.
Therefore, if a safe RL algorithm penalizes too hard on cost return $V_c^\pi$, its trajectory will end up at a point on the boundary that has a low reward.
The fast drop in cost during training usually leads to a convergence point $\pi_\text{Safe}$ that is far away from the optimal policy $\pi^*$.
This aligns with our observation in experiments where safe RL agents often get stuck in a position surrounded by hazards and cannot find a way out.

In this case, offline policy extraction from expert demonstrations can provide a head start, which boosts the original initial $\pi_0$ to $\pi'_0$ (closer to $\pi^*$) in Fig.~\ref{fig:reward-cost}(b).
The online distillation can start from $\pi'_0$ that skips exploring the environment from scratch and thus requires less effort to $\pi_\text{GOLD}$ of higher quality than $\pi_\text{Safe}$ even with the same online training RL backbone. Thus, we propose a new training scheme that pushes the trajectory toward the optimal policy $\pi^*$ by leveraging demonstrations to extract prior skills and perform online safe RL finetuning.

\section{Guided Online Distillation}

In this section, we present our proposed method: Guided Online Distillation (GOLD).
It consists of two stages: 1) extracting a large-scale guide policy from offline demonstration, and 2) distilling a robust but lightweight policy through online exploration with the guidance of the guide policy.

\subsection{Extracting Expert from Offline Demonstration}
\label{sec:decision-transformer}

Offline policy training from demonstration has been a popular research topic, and many methods have been proposed. 
DT~\cite{chen2021decision} is an approach that lies in between BC and offline RL and proves to be competent.
It adopts a similar loss function and training scheme as BC but also considers reward signals as offline RL.
We therefore choose to apply DT to extract expert policies from offline demonstration and empirically show that it is superior to both BC and offline RL for safety-critical navigation and autonomous driving tasks.

The trajectory representation and model architecture follow the design in~\cite{chen2021decision}. 
Specifically, we choose to represent the trajectory by three modalities: observation, action, and returns-to-go.
Formally, the trajectory representation is
$\tau = \left( \hat{R}_1, \vs_1, \va_1, \dots, \hat{R}_T, \vs_T, \va_T \right)$,
where
$\hat{R}_t = \sum_{i=t}^T r_i$ is the returns-to-go, 
$\vs_t$ and $\va_t$ are the observation and the action at time $t$.
The model is fed with the most recent $K$ timesteps, encompassing a total of $3K$ tokens.
In the experiments in this paper, we find the default setting of $K=20$ to be suitable for most of the tasks.
A GPT~\cite{radford2018improving} model processes the inputs by autoregressive modeling.
Leveraging a dataset of offline trajectories, we extract minibatches with a sequence length of $K$ from the dataset.
The prediction head linked to the input token $o_t$ is trained to predict action $a_t$.
The loss is only evaluated on the predicted action, as no performance gain is reported by predicting observation and returns-to-go~\cite{chen2021decision}.
In our case, the loss is defined to be
\begin{gather}
    L_{DT} = || \va - \hat{\va} ||^2
\end{gather}
where $\va$ is the ground truth action, and $\hat{\va}$ is the predicted action by the DT.

\subsection{Online Policy Distillation}
\label{sec:online-policy-distillation}

DT for expert policy extracting from offline demonstration improves the performance, but it sacrifices computation efficiency and robustness.
Transformers are bulk in size, so they consume large amount of computation resources.
This can be critical when deploying them as decision-making modules on real systems that request fast response frequency. The offline demonstration is also not comprehensive, and hence there are always hazardous corner cases not included. It results in a guide policy that is only reliable close to the in-distribution areas within the offline data support. Therefore, we propose to distill a lightweight policy network and improve its robustness to out-of-distribution hazardous areas with the guidance of DT during online exploration within the task environment.

The backbone of online distillation is based on JSRL~\cite{uchendu2023jump}.
We define a guide policy as the pre-trained DT from Sec.~\ref{sec:decision-transformer}, whose parameters are frozen during online distillation.
The policy network to be distilled is designed to be a lightweight Multi-Layer Perceptron (MLP).
On the one hand, JSRL makes sure the reward maintains its stable improvements, instead of dropping dramatically in naive online fine-tuning methods.
The skills of the guide DT are distilled into the lightweight policy by exposing it to a high-reward trajectory distribution.
On the other hand, the states induced by the guide policy are also relatively safe, where the agent explores to learn fine-grained information on the costs. 
This makes the lightweight exploration policy not only training efficient but also robust.

However, storing the rollouts of both guide and lightweight policies in one replay buffer causes a mixed data distribution, which induces problems for the critic learning in RL algorithms.
Actor-critic methods aim to approximate an optimal Q-function corresponding to the current parameterized policy $\pi(\va|\vs)$, which satisfies the equation
\begin{gather*}
\begin{aligned}
    &Q^\pi (\vs_t, \va_t) = r(\vs_t, \va_t) \: + \\
    &\gamma \E_{\vs_{t+1} \sim T(\vs_{t+1} | \vs_t, \va_t), \va_{t+1} \sim \pi (\va_{t+1} | \vs_{t+1})}\left[ Q^\pi (\vs_{t+1}, \va_{t+1}) \right].
\end{aligned}
\end{gather*}
The Q-function should be evaluating the future cumulative reward under the data distribution induced by the current policy. 
If the training data is collected by multiple policies as the online distillation process, the mixed and skewed data distribution will cause the Q value prediction to be inaccurate on trajectories collected by the current lightweight policy being trained.

We propose to resolve the aforementioned problem by leveraging Implicit Q Learning (IQL)~\cite{kostrikov2021offline} as the training algorithm during online distillation.
IQL approximates a Q-function without an explicit policy by expectile regression.
Specifically, it first estimates expectiles only with respect to the actions in the support of the data by first approximating a value function $V_\psi(\vs)$ with a loss $L_V(\psi)$,
\begin{gather*}
    L_V(\psi) = \E_{(\vs, \va) \sim \mathcal{D}} \left[ L_2^\tau (Q_{\hat{\theta}} (\vs, \va) - V_\psi (\vs)) \right],
\end{gather*}
where $L_2^\tau (u) = |\tau - \mathds{1}(u<0)| u^2$, and $\tau$ is the expectile.
It then avoids injecting stochasticity from the state distribution by averaging over the stochasticity from the dynamics transitions and fitting a Q-function $Q_\theta$ with a loss $L_Q(\theta)$,
\begin{gather*}
    L_Q(\theta) = \E_{(\vs, \va, \vs') \sim \mathcal{D}} \left[ r(\vs, \va) + \gamma V_\psi (\vs') - Q_\theta (\vs, \va) \right]^2.
\end{gather*}
The fitted Q-function corresponds to the upper expectile of the returns, which makes it approximate better the Q-function corresponding to the optimal policy.
This decoupling between the Q-function approximation and the current policy is suitable for GOLD.
The Q-function is not sensitive to the mixed state trajectory distribution in the replay buffer anymore, instead, it corresponds to the optimal policy.

\subsection{Practical Implementation}

\begin{algorithm}[t]
\SetAlgoLined
    \textbf{Initialize}: A decision transformer (DT) $\pi_{\mu}^g$ for guide policy, an lightweight policy network $\pi_\varphi$, a Q-network $Q_\theta$, A target network $Q_{\Bar{\theta}} = Q_\theta$,  a training dataset $\mathcal{D}$, a replay buffer $\mathcal{B}$\;
    \textbf{\slash \slash \space Prior skills extraction from offline demonstration} \\
    \For{step $n$ in range($0$, $N$)}{
        Sample a batch of $b$ trajectory segments $\tau_{t-H}^t$ from the dataset $\mathcal{D}$\;
        Update $\mu$: $\mu_n \leftarrow \mu_{n-1} + \epsilon_\mu \nabla_\mu L_{DT}$
    }
    \textbf{\slash \slash \space Online distillation procedure} \\
    \For{guide step $h$ in $[H_1, H_2, \dots, H_m]$}{
        Assign a non-stationary policy $\pi$ defined at each timestep: $\pi_{1:h} = \pi_\mu^g$, $\pi_{h+1:H} = \pi_\varphi$\;
        Collect rollouts by $\pi$ and append them to the replay buffer $\mathcal{B}$\;
        \For{train step $m$ in range($0$, $M$)}{
            Sample a batch $(\vs_t, \va_t, r_t, \vs_{t+1})$ from $\mathcal{B}$\;
            Update $Q_\theta$ and $\pi_\varphi$ by IQL\;
        }
    }
    \caption{Training Procedure of GOLD}
    \label{alg:train}
\end{algorithm}

We summarize the complete proposed algorithm GOLD in Algo.~\ref{alg:train}.
A DT is first trained from offline demonstration, which later serves as the guide policy during online distillation.
A lightweight exploration policy network is then trained interactively in the task environment by IQL.
In GOLD, the safety constraints are enforced by reward shaping, adapting IQL to safe RL settings, i.e., its actual reward is a linear combination of the original reward and cost 
$$\text{reward}_\text{new} = \text{reward} + \lambda \cdot \text{cost},$$
since we are focused on safety-critical tasks in this paper.

\section{Experiments}
\begin{table*}[t]
    \centering
    \begin{adjustbox}{max width=0.95\textwidth}
        \begin{tabular}{l|c|c|c||c|c|c|c|c}
        \toprule
        \multicolumn{2}{c|}{}& \multicolumn{2}{c||}{\textbf{Offline}} & \multicolumn{5}{c}{\textbf{Online}} \\
        \midrule
        \multicolumn{2}{c|}{\textbf{Task}} & \textbf{BC} & \textbf{DT} & \textbf{IQL} & \textbf{CVPO} & \textbf{GOLD (BC-IQL)} & \textbf{GOLD (DT-CVPO)} & \textbf{GOLD (DT-IQL)} \\
        \midrule 
        \midrule
        \multirow{2}{6em}{\parbox{1.6cm}{\centering \textbf{Car-Circle}}} & r $\uparrow$
            & $366.9 \pm 10.4$
            & $\mathbf{450.3 \pm 53.8}$
            & $630.7 \pm 26.4$
            & $502.5 \pm 10.8$
            & $628.4 \pm 20.6$
            & $613.7 \pm 26.3$
            & $\mathbf{688.3 \pm 4.2}$
            \\
            
            & c $\downarrow$
            & $41.4 \pm 5.3$
            & $\mathbf{40.4 \pm 6.3}$
            & $17.5 \pm 2.8 $
            & $7.4 \pm 3.3$
            & $13.6 \pm 1.4$
            & $3.9 \pm 0.5$
            & $\mathbf{3.2 \pm 1.5}$
            \\
        \midrule
        \multirow{2}{6em}{\parbox{1.6cm}{\centering \textbf{Car-Gather}}} & r $\uparrow$
            & $5.6 \pm 2.31 $ 
            & $\mathbf{7.1 \pm 1.52}$
            & $10.3 \pm 1.22 $
            & $10.2 \pm 0.87 $
            & $12.8 \pm 2.63 $
            & $11.9 \pm 0.44 $
            & $\mathbf{14.0 \pm 1.98}$ 
            \\
            \cmidrule{2-9}
            & c $\downarrow$
            & $0.42 \pm 0.17$
            & $\mathbf{0.38 \pm 0.16}$
            & $0.23 \pm 0.12$
            & $0.18 \pm 0.04$
            & $0.19 \pm 0.27$
            & $0.15 \pm 0.02$
            & $\mathbf{0.14 \pm 0.04}$
            \\
        \midrule
        \multirow{2}{6em}{\parbox{1.6cm}{\centering \textbf{Point-Goal}}} & r $\uparrow$
            & $19.2 \pm 1.4$
            & $\mathbf{24.1 \pm 0.5}$
            & $31.6 \pm 3.2$
            & $32.1 \pm 5.3$
            & $32.1 \pm 5.8$
            & $31.4 \pm 3.8$
            & $\mathbf{33.9 \pm 6.5}$
            \\
            \cmidrule{2-9}
            & c $\downarrow$
            & $16.7 \pm 2.4$
            & $\mathbf{15.2 \pm 4.6}$
            & $10.5 \pm 3.9$
            & $8.5 \pm 1.4$
            & $\mathbf{8.0 \pm 3.8}$
            & $11.5 \pm 2.9$
            & $8.3 \pm 1.3$
            \\
        \midrule
        \multirow{2}{6em}{\parbox{1.6cm}{\centering \textbf{Point-Button}}} & r $\uparrow$
            & $23.7 \pm 5.2 $
            & $\mathbf{27.8 \pm 4.2}$
            & $35.1 \pm 4.7 $
            & $38.2 \pm 2.4 $
            & $41.1 \pm 3.1 $
            & $39.2 \pm 2.4 $
            & $\mathbf{44.8 \pm 3.1}$
            \\
            \cmidrule{2-9}
            & c $\downarrow$
            & $18.5 \pm 5.9$
            & $\mathbf{17.3 \pm 2.5}$
            & $8.4 \pm 2.1$
            & $7.5 \pm 2.8$
            & $\mathbf{6.2 \pm 4.2}$
            & $6.8 \pm 2.9$
            & $6.5 \pm 1.1 $
            \\
        \midrule
       \multirow{2}{6em}{\parbox{1.6cm}{\centering \textbf{Point-Push}}} & r $\uparrow$
            & $2.1 \pm 3.2 $
            & $\mathbf{4.6 \pm 2.1}$
            & $5.3 \pm 1.0 $
            & $2.5 \pm 0.3 $
            & $6.6 \pm 1.6 $
            & $4.2 \pm 1.1 $
            & $\mathbf{8.0 \pm 2.3}$
            \\
            \cmidrule{2-9}
            & c $\downarrow$
            & $50.2 \pm 8.1$
            & $\mathbf{45.4 \pm 5.8}$
            & $34.1 \pm 9.1$
            & $19.3 \pm 4.2$
            & $24.3 \pm 3.8$
            & $\mathbf{18.3 \pm 4.5}$
            & $20.4 \pm 6.9$
            \\
        \midrule
        \multirow{2}{6em}{\parbox{1.6cm}{\centering \textbf{Car-Goal}}} & r $\uparrow$
            & $13.2 \pm 1.7 $
            & $\mathbf{16.4 \pm 3.4}$
            & $22.8 \pm 1.3 $
            & $19.8 \pm 1.9 $
            & $27.9 \pm 2.1 $
            & $28.4 \pm 1.2 $
            & $\mathbf{30.5 \pm 5.4}$
            \\
            \cmidrule{2-9}
            & c $\downarrow$
            & $53.4 \pm 2.1$
            & $\mathbf{49.9 \pm 6.3}$
            & $20.4 \pm 5.2$
            & $24.3 \pm 6.6$
            & $14.6 \pm 2.6$
            & $12.2 \pm 4.1$
            & $\mathbf{10.8 \pm 3.2}$
            \\
        \midrule
        \multirow{2}{6em}{\parbox{1.6cm}{\centering \textbf{Car-Button}}} & r $\uparrow$
            & $19.6 \pm 2.2$
            & $\mathbf{26.5 \pm 3.7}$
            & $28.9 \pm 10.4 $
            & $27.1 \pm 3.8 $
            & $36.1 \pm 3.0 $
            & $30.5 \pm 3.8 $
            & $\mathbf{42.0 \pm 2.6}$
            \\
            \cmidrule{2-9}
            & c $\downarrow$
            & $34.1 \pm 10.5$
            & $\mathbf{20.8 \pm 9.3}$
            & $12.5 \pm 6.8$
            & $8.8 \pm 1.3$
            & $9.8 \pm 5.0$
            & $7.1 \pm 4.3$
            & $\mathbf{6.5 \pm 6.1}$
            \\
        \midrule
        \multirow{2}{6em}{\parbox{1.6cm}{\centering \textbf{Car-Push}}} & r $\uparrow$
            & $1.5 \pm 0.1 $
            & $\mathbf{2.6 \pm 0.2}$
            & $3.8 \pm 1.1 $
            & $3.0 \pm 0.4 $
            & $4.5 \pm 0.2 $
            & $4.2 \pm 0.2$ 
            & $\mathbf{5.1 \pm 0.4}$
            \\
            \cmidrule{2-9}
            & c $\downarrow$
            & $65.3 \pm 9.8$
            & $\mathbf{58.9 \pm 11.6}$
            & $23.3 \pm 2.9$
            & $19.3 \pm 5.8$
            & $19.4 \pm 1.9$
            & $\mathbf{15.7 \pm 2.3}$
            & $17.4 \pm 1.1$
            \\
        \midrule
        \midrule
        \multirow{4}{6em}{\parbox{1.6cm}{\centering \textbf{MetaDrive \\ Waymo}}} & r $\uparrow$
            & $115.78 \pm 132.89$
            & $\mathbf{133.48 \pm 190.50}$
            & $26.29 \pm 68.82$
            & $113.48 \pm 163.84$
            & $141.93 \pm 189.54$
            & $115.66 \pm 163.18$
            & $\mathbf{143.69 \pm 175.98}$
            \\
            \cmidrule{2-9}
            & c $\downarrow$
            & $ \mathbf{ 1.14 \pm 1.96}$
            & $1.25 \pm 1.92$
            & $ 2.57 \pm 2.36 $
            & $ 1.05 \pm 1.84$
            & $ \mathbf{ 1.03 \pm 1.85 }$
            & $ 1.25 \pm 2.07$
            & $1.15 \pm 2.05$
            \\
            \cmidrule{2-9}
            & sr $\uparrow$
            & $53\%$
            & $\mathbf{54\%}$
            & $40\%$
            & $58\%$
            & $63\%$
            & $62\%$
            & $\mathbf{73\%}$
            \\
        \bottomrule
        \end{tabular}
    \end{adjustbox}
    \caption{
        The performance of offline policy extraction and online policy distillation.
        Metric notations are defined as, r: reward, c: cost.
        For the realistic driving environment based on WOMD, we also compare success rate as sr.
    }
    \label{tab:full-results}
\end{table*}

\subsection{Experiment Setting}

\subsubsection{Safety Gym \& Bullet Safety Gym}

\texttt{safety-gym}~\cite{Ray2019} and \texttt{bullet-safety-gym}~\cite{gronauer2022bullet} are open-source frameworks which is designed to train and evaluate safety performance across many tasks and environments, distinct in complexity and design.
The observation of an agent is set to include the agent's own body state, the sensing information on obstacles given by pseudo laser rays, and task-specific information such as distance to goals.
We pick five tasks with two different agent types.
The tasks include \texttt{Circle} and \texttt{Gather}, \texttt{Goal}, \texttt{Button}, \texttt{Push}, and the agent types are \texttt{Point} and \texttt{Car}.
We perform training and evaluation in different combinations of tasks and agents.

\subsubsection{MetaDrive}

a lightweight yet powerful driving simulator~\cite{li2022metadrive}, which provides convenient scene composition with various road maps and traffic settings that are critical for generalizable RL.
The simulation is realistic as it leverages an accurate physical engine and emulates sensory input.
The driving scenes can be replayed from real-world traffic data such as WOMD~\cite{Ettinger_2021_ICCV, Kan_2023_arxiv}, NuScenes~\cite{caesar2020nuscenes}, Argoverse~\cite{chang2019argoverse}, etc.
The observation consists of pseudo Lidar-like cloud points, navigation information represented by waypoints, and ego states, including steering, heading, velocity, and relative distance to boundaries.
The action space contains the acceleration and steering of the ego vehicle.

\subsubsection{Offline Datasets}

In our problem setting, we assume access to offline datasets collected from expert demonstration. 
These demonstrations are near-optimal, which contain few safety-critical situations, but mostly trajectories with high reward returns. 
Formally, the dataset contain $N$ expert trajectories, each of which is represented by $H$ tuples $\{ (\vs_t^k, \va_t^k, \vs_{t+1}^k, r_t^k)_{t=0}^T \}_{k=1}^N$, where $t \in [0, T]$ is the time step from $0$ to $H$, and $k \in [1, N]$ is the episode number from $1$ to $N$. In the \texttt{(bullet)-safety-gym} environments, the offline datasets are collected by expert RL policies to imitate human experts, which are trained in online settings by SAC with carefully tuned reward shaping for data collection purposes. 
These expert RL policies are able to reach high reward returns and satisfy cost requirements.
We use Waymo Open Datasets as the offline demonstration dataset for the MetaDrive task.
These datasets are recordings of real traffic scenes, which are generated by human drivers.

\subsection{Baselines}

We compare our proposed GOLD to its own variants and state-of-the-art safe RL methods, including:
\begin{itemize}
    \item \textbf{Safe RL Trained from Scratch}: To show the role of the guide policy in GOLD, we choose safe RL methods: Implicit Q Learning (IQL)~\cite{kostrikov2021offline} equipped with reward shaping and Constrained Variational Policy Optimization (CVPO)~\cite{liu2022constrained} as baselines, which are both trained from scratch.
    They are not warm-started by the guide policy (DT), since we intend to keep the number of parameters and network structure the same as GOLD.
    \item \textbf{Variants of the proposed method}: We also demonstrate how different components in the proposed method contribute to the final performance. 
    For the guide policy trained from offline datasets, we compare DT with BC.
    For the RL backbone of online distillation and training, we compare IQL with CVPO.
    In summary, we have four variants, namely, GOLD (BC-IQL), GOLD (DT-CVPO), and GOLD (DT-IQL).
\end{itemize}

\begin{figure}[t]
    \centering
    \includegraphics[width=0.49\textwidth]{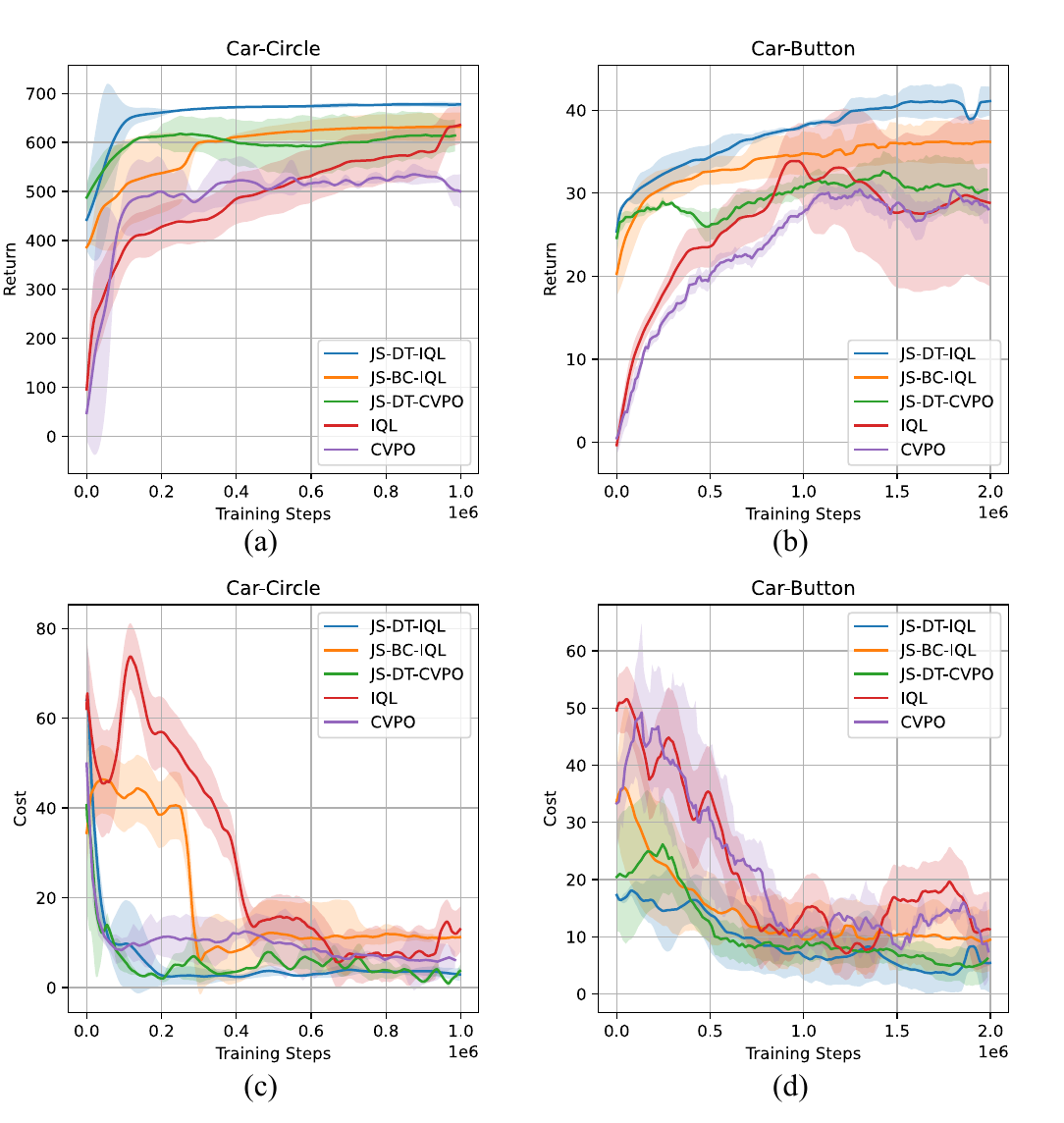}
    \caption{
        The learning curves of GOLD (DT-IQL) and baselines.
        (a)(c): The reward and cost curves in Car-Circle.
        (b)(d): The reward and cost curves in Car-Button.
    }
    \label{fig:learning-curves}
\end{figure}

\subsection{Guide Policy Performance}

The performance of various offline policy extraction methods in terms of both reward and cost is listed in Tab.~\ref{tab:full-results}.
The columns of offline methods show that DTs are superior in both reward and cost than BC in all benchmark tasks. In convention, BC or RL adopts MLP as the policy or value network.
However, DTs use large models such as pre-trained GPT2 as the network backbone.
This difference dramatically increases the model capacity of the policy network and thus improves the final performance by a large margin.

We observe the performance of DT is correlated with the offline dataset size.
Typically, it benefits from enlarging the size of the dataset.
We find it is sufficient to show the difference between DT and MLP using datasets of a scale of $100$k trajectories for \texttt{(bullet-)safety-gym} tasks, and $10$k trajectories from WOMD for Metadrive tasks.
Due to limited hardware accelerator resources, we cannot perform more computation-intensive guide policy extraction on larger datasets.
We leave it to future work as it pertains to the current trend of leveraging richer and bulkier datasets.

\subsection{Online Distillation Evaluation}
\label{sec:benchmark-perf}
With the guide policy extracted from offline demonstrations, our proposed method finetunes and distills a lightweight yet more powerful policy network through interactions within online environments.

\subsubsection{Computation Efficiency}
The online training distills a much smaller policy network, i.e. a two-layer MLP with a hidden size of $256$, which is standard in most RL problem settings.
The number of parameters of the MLP is negligible compared to the huge transformer used by the guide policy, which usually has $10$x times more parameters ($670$k in \texttt{safety-gym} tasks).
The computation efficiency is thus noticeable and becomes an advantage when deploying these MLP policies compared to the huge DT, if the performance is above threshold.
For the benchmark tasks in our setting, the online distilled MLP runs at $0.03$s per 100 inference runs, compared to $0.31$s per 100 inference runs of DT on a single NVIDIA GeForce RTX 2080 Ti GPU.

\subsubsection{Policy Performance}
The performance of all baselines and variants are listed in Tab.~\ref{tab:full-results} under the tab ``Online''. 
Our proposed method outperforms all baselines in terms of reward and is superior in most tasks in terms of cost. We can see that algorithms with a guide in online distillation, i.e. variants of GOLD, perform better than training from scratch, i.e., IQL and CVPO.
Regular safe RL algorithms tend to get stuck with local optimal solutions because they are discouraged from high-risk exploration in the environment by their cost constraints.
However, with the guidance of pre-trained offline policies, GOLD avoids the exploration that leads to many failures before success and can discover highly lucrative solutions. Fig.~\ref{fig:learning-curves} shows that the agent learns faster and better when equipped with a guide policy.

\begin{figure}[t]
    \centering
    \includegraphics[width=0.47\textwidth]{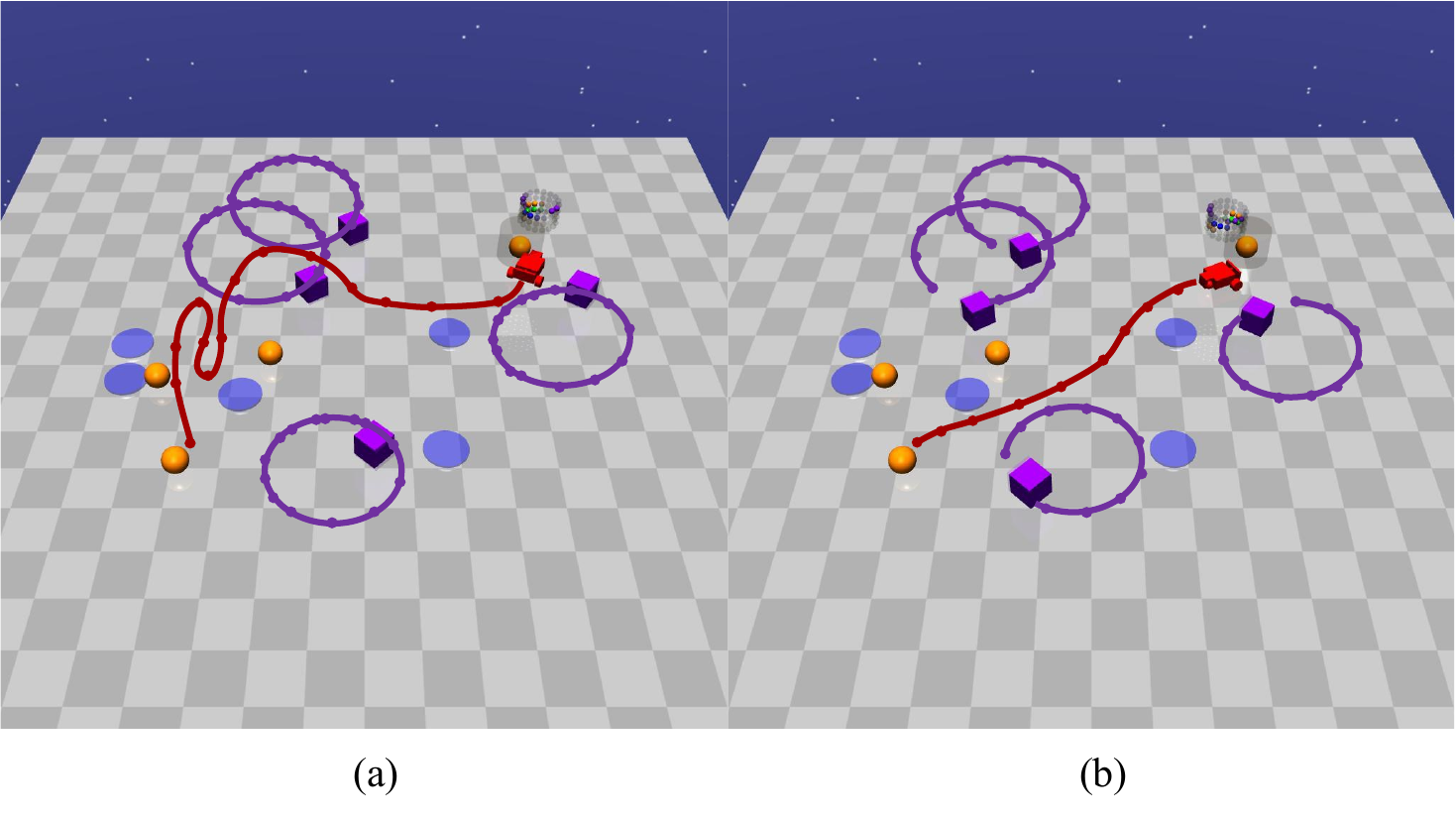}
    \caption{
        Sampled trajectories in the Car-Button task.
        (a): GOLD (BC-IQL). The red ego car avoids the moving purple hazardous obstacle at first and struggles to find its way in the later stage.
        (b): GOLD (DT-IQL). The ego car can find an efficient trajectory to reach the goal, thanks to the prior knowledge inherited in the guide DT policy.
    }
    \label{fig:traj}
\end{figure}

We also show the advantage of guidance during online distillation in Fig.~\ref{fig:traj}.
Here, the red car intends to press the yellow button on the upper right corner starting from the lower left corner, without touching the purple and blue hazardous obstacles.
The red curve behind the car is its historical trajectory.
The purple obstacles are moving, while the blue obstacles are staying still.
In Fig.~\ref{fig:traj}(a), the ego car trained with CVPO only finds a local optimal solution and chooses to avoid the bottom purple box at the beginning, which results in zigzagging trajectory and hence lower reward in the later stage of the episode.
In contrast, the ego car trained with our method learns to find the most direct and safe way to the goal by the guidance of the expert policy in the early training stage and thus obtains higher reward and lower cost.

In Tab.~\ref{tab:full-results}, we also show that the online distillation performance improves with a better guide policy. The algorithms GOLD (DT-*) usually obtain better rewards compared to GOLD (BC-IQL).
This aligns with the intuition that a better teacher reduces the effort to learn the same level of skills.

Plus, GOLD (DT-IQL) typically performs better than GOLD (DT-CVPO), as shown in Tab.~\ref{tab:full-results} and Fig.~\ref{fig:learning-curves}, even though they both adopt DT as the guide policy.
This is because of the data distribution mismatch in the replay buffer, which is mentioned in Sec.~\ref{sec:online-policy-distillation}.
CVPO learns Q-functions evaluating only the current explore policy being trained, which mismatches the trajectories collected by a mixture of guide and explore policies.
Using IQL as the online finetuning and distillation backbone solves this problem because it learns Q-functions only by evaluating the optimal policy in the environment, which in theory, can learn from data collected by any policy.

\subsection{Realistic Experiments in Driving Scenarios}

Our method is applicable to and effective in realistic scenarios, which we demonstrate by experiments on MetaDrive.
These experiments are fairly close to real-world scenes because we make MetaDrive replay vehicle trajectories from WOMD. The observations input to the ego agent, including Lidar cloud points, navigation information, and ego states, also resemble the real-world setting.
The goal of the ego vehicle is to arrive at a specific target position defined in WOMD.
We randomly choose $10$k scenarios from WOMD for training and $1$k scenes for testing.

As shown in Tab.~\ref{tab:full-results}, our method surpasses baselines by around $15\%$ in reward and maintains the cost below the threshold.
The success rate is increased by $12\%$ compared to the best variant, which shows GOLD is capable in realistic driving tasks.
The evaluation results and analysis on previous benchmarks in Sec.~\ref{sec:benchmark-perf} are transferrable to realistic tasks, which confirms the performance of GOLD is correlated to the guide policy quality.
This supports and justifies our design of offline policy extraction by DT.

The driving experiments further validate that bringing in prior skills during online distillation is necessary for learning high-quality policy in real-world safety-critical scenarios. 
CVPO or IQL from scratch is too conservative to explore because it is almost impossible to discover useful skills without severe cost violations.
GOLD skips the risky exploration in this safety-critical environment.
With the offline trained DT as guidance, GOLD can distill and improve a lightweight policy network without struggling to jump out of the most hazardous areas.
Also, CVPO outperformed IQL by a large margin when trained from scratch, but GOLD (DT-IQL) surpasses GOLD (DT-CVPO). This confirms that IQL's decoupling of Q-function and policy training works seamlessly with GOLD.
More video demonstrations can be found on \url{https://sites.google.com/view/guided-online-distillation}.

\section{Conclusion}

We propose a new offline-to-online training scheme named Guided Online Distillation for safety-critical tasks.
A large-scale guide policy is first extracted from offline demonstrations. It serves as a guide for online distillation, where a lightweight policy is distilled through interactions with the task environment.
This lightweight network can meet computation speed requirements in realistic settings, in contrast to the bulk guide policy.
The guided distillation saves the policy from being repeatedly exposed to hazards during its exploration to find useful skills, which improves its training efficiency and final performance.
Experiments in both benchmarks and real-world driving experiments based on the WOMD show that the distilled policy by GOLD surpasses safe RL baselines that are trained from scratch.

\newpage
\bibliographystyle{ieeetr}
\bibliography{reference.bib}

\end{document}

%% file: math_commands.tex
\usepackage{amsmath,amsfonts,bm}

\def\eqref#1{equation~\ref{#1}}

\def\1{\bm{1}}

\def\va{{\bm{a}}}

\def\vs{{\bm{s}}}

\DeclareMathAlphabet{\mathsfit}{\encodingdefault}{\sfdefault}{m}{sl}
\SetMathAlphabet{\mathsfit}{bold}{\encodingdefault}{\sfdefault}{bx}{n}

\newcommand{\E}{\mathbb{E}}

\newcommand{\R}{\mathbb{R}}

